# Cut-and-Paste with Precision: a Content and Perspective-aware Data Augmentation for Road Damage Detection

Punnawat Siripathitti[a,b], Florent Forest[b], Olga Fink[b]

[a] *KTH, Stockholm, Sweden*
[b] *Intelligent Maintenance and Operations Systems, EPFL, Lausanne, Switzerland*

---

Damage to road pavement can develop into cracks, potholes, spallings, and other issues posing significant challenges to the integrity, safety, and durability of the road structure. Detecting and monitoring the evolution of these damages is crucial for maintaining the condition and structural health of road infrastructure. The condition monitoring of road infrastructure can be automated via image-based visual road damage detection. In recent years, researchers have explored various data-driven methods for image-based damage detection in road monitoring applications (Cao et al., 2020). The field gained attention with the introduction of the Road Damage Detection Challenge (RDDC2018), encouraging competition in developing object detectors on street-view images from various countries. Leading teams have demonstrated the effectiveness of ensemble models, mostly based on the YOLO and Faster R-CNN series (Arya et al., 2022). Ensemble learning is a powerful technique for improving the generalization of object detection models, but comes at the expense of being time-consuming and resource-intensive. Data augmentations have also shown benefits in object detection within the computer vision field, including transformations such as random flipping, cropping, color jittering, grayscaling, mosaic, cutting out patches (Zhong et al., 2017), as well as cut-and-pasting object instances (Dwibedi et al., 2017). Applying cut-and-paste augmentation to road damages appears to be a promising approach to increase data diversity. However, the standard cut-and-paste technique, which involves sampling an object instance from a random image and pasting it at a random location onto the target image, has demonstrated limited effectiveness for road damage detection (Pham et al., 2020). This method overlooks the location of the road and disregards the difference in perspective between the sampled damage and the target image, resulting in unrealistic augmented images. The limitation highlights the need for more sophisticated approaches.

In this work, we propose an improved Cut-and-Paste augmentation technique that is both content-aware (i.e. considers the true location of the road in the image) and perspective-aware (i.e. takes into account the difference in perspective between the injected damage and the target image). For content-awareness, we leverage a pretrained semantic segmentation model (Cheng et al., 2019) to generate binary road masks, constraining the location of the injected damages. Furthermore, perspective is affected by the pitch angle of the camera and the distance of the damage to the camera (see Fig. 1). We achieve perspective-awareness by transforming the road mask into a perspective map (Lis et al., 2022). Then, images are grouped into bins according to their pitch angle. Damage instances with a similar perspective range are then sampled and injected onto the road surface of the target image, applying a perspective transform similar to the one proposed by (Zhou et al., 2022). The sampling probability is uniform within each bin to preserve the original data distribution. Additionally, we generate heatmaps of instance locations for each pitch angle range. Interestingly, if the images were taken from a country where cars are driving on the left side of the road, the heatmap distribution is concentrated on the right side of the image. We use the heatmaps to increase the probability of injecting damages into this region, and parts of the damage placed outside the road mask are cropped out. Finally, we also use Poisson blending (Pérez et al., 2003) to better blend the damage into the background. The augmentation process is illustrated in Fig. 1.

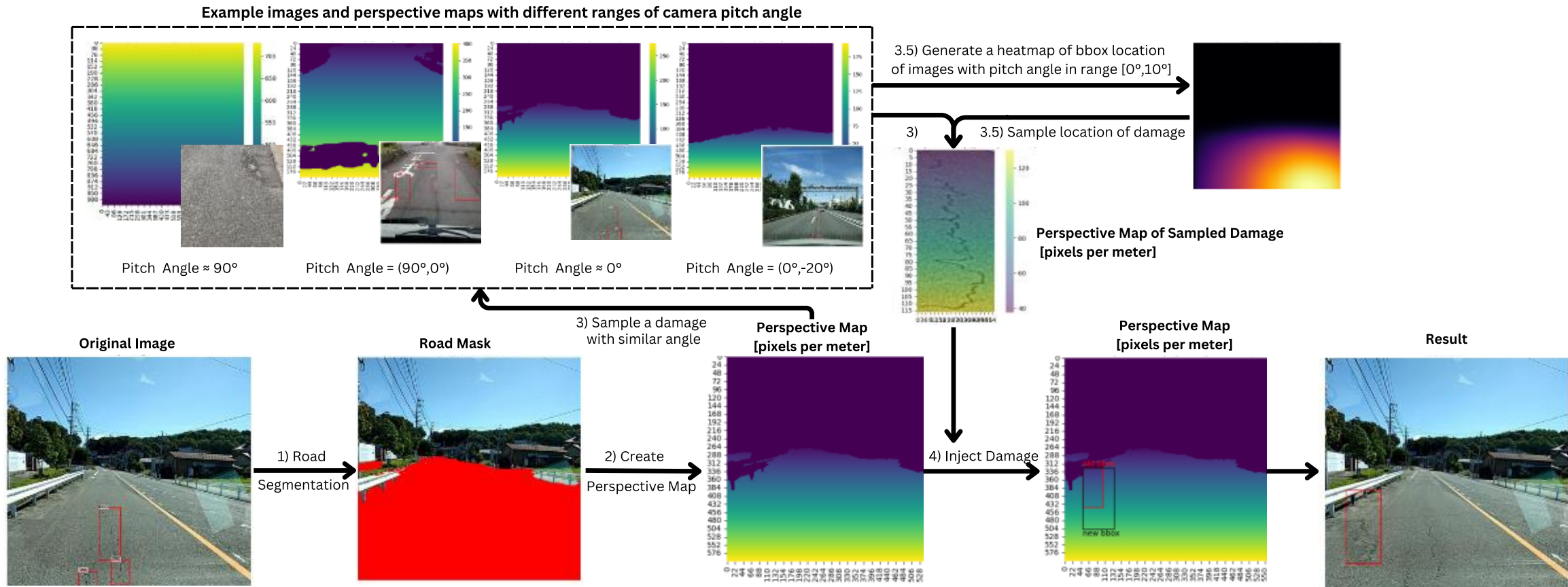

Fig.1. Diagram of our proposed Content and Perspective-Aware damage injection process.

We evaluate our method on the Japan subset of the RDDC2022 dataset (Arya et al., 2022). Annotations consist of four damage classes: Longitudinal Crack (D00), Transverse Crack (D10), Alligator Crack (D20) and Pothole (D40). We divide the dataset following a 90/10 train/test split. The applied object detection model is Faster R-CNN with the X-101-FPN backbone, implemented in Detectron2. Performance evaluation on the test set, as reported in Tab. 1, includes mean Average Precision at 50% IoU (mAP50) and AP for each damage class. The proposed method demonstrates a performance improvement over the baseline model, particularly in enhancing the detection accuracy for damage classes D00, D20, and D40. It also showcases a lower loss, indicating better generalization. These results highlight the effectiveness of our content-aware and perspective-aware cut-and-paste augmentation in improving the precision of road damage detection models.

Table 1. Road damage detection results with different data augmentation strategies (RDDC2022 dataset – Japan).

| Method | Damage Injection | Content Awareness | Perspective Awareness | Damage Class (AP) | | | | mAP50 | Test loss |
|---|---|---|---|---|---|---|---|---|---|
| | | | | D00 | D10 | D20 | D40 | | |
| Baseline | ✗ | ✗ | ✗ | 22.84 | **18.87** | 33.00 | 21.58 | 56.52 | 0.86 |
| Cut-and-paste | ✓ | ✗ | ✗ | 21.82 | 17.22 | 34.35 | 21.33 | 56.95 | 0.83 |
| | ✓ | ✓ | ✗ | 22.94 | 16.59 | 34.73 | 21.37 | 56.55 | 0.82 |
| Ours | ✓ | ✓ | ✓ | **23.33** | 17.33 | **35.13** | **24.29** | **58.19** | **0.81** |